\documentclass[letterpaper, 10 pt, conference]{ieeeconf}  % Comment this line out if you need a4paper

\IEEEoverridecommandlockouts                              % This command is only needed if 
\usepackage{amsmath} % assumes amsmath package installed
\usepackage{amssymb}  % assumes amsmath package installed
\usepackage{balance}
\usepackage{xcolor}
\usepackage{graphicx} 
\usepackage{multicol,caption}% http://ctan.org/pkg/multicols
\usepackage{multirow}
\usepackage{multicol}
\usepackage{subfigure}
\usepackage{url} 

\title{\LARGE \bf
Human-Robot Collaborative Minimum Time Search through Sub-priors in Ant Colony Optimization
}

\author{Oscar Gil$^{\dagger}$ and Alberto Sanfeliu$^{\dagger}$% <-this % stops a space
\thanks{Work supported under the European project CANOPIES with grant number H2020- ICT-2020-2-101016906 and JST Moonshot R \& D Grant Number JPMJMS2011-85.}% <-this % stops a space
\thanks{$^{\dagger}$The authors are with the Institut de Robòtica i Informàtica Industrial (CSIC-UPC),
        Llorens Artigas 4-6, 08028 Barcelona, Spain.
        {\tt\small \{ogil, sanfeliu\}@iri.upc.edu}}%
}

\begin{document}

\maketitle
\thispagestyle{empty}
\pagestyle{empty}

%%%%%%%%%%%%%%%%%%%%%%%%%%%%%%%%%%%%%%%%%%%%%%%%%%%%%%%%%%%%%%%%%%%%%%%%%%%%%%%%
\begin{abstract}

Human-Robot Collaboration (HRC) has evolved into a highly promising issue owing to the latest breakthroughs in Artificial Intelligence (AI) and Human-Robot Interaction (HRI), among other reasons. This emerging growth increases the need to design multi-agent algorithms that can manage also human preferences. This paper presents an extension of the Ant Colony Optimization (ACO) meta-heuristic to solve the Minimum Time Search (MTS) task, in the case where humans and robots perform an object searching task together. The proposed model consists of two main blocks. The first one is a convolutional neural network (CNN) that provides the prior probabilities about where an object may be from a segmented image. The second one is the Sub-prior MTS-ACO algorithm (SP-MTS-ACO), which takes as inputs the prior probabilities and the particular search preferences of the agents in different sub-priors to generate search plans for all agents. The model has been tested in real experiments for the joint search of an object through a Vizanti web-based visualization in a tablet computer. The designed interface allows the communication between a human and our humanoid robot named IVO. The obtained results show an improvement in the search perception of the users without loss of efficiency.

\end{abstract}

%%%%%%%%%%%%%%%%%%%%%%%%%%%%%%%%%%%%%%%%%%%%%%%%%%%%%%%%%%%%%%%%%%%%%%%%%%%%%%%%
\section{INTRODUCTION}

For thousands of years, mankind has relied on collaboration between individuals to perform tasks as optimally as possible in a wide variety of situations. Actually, the increasing use of robots in a wide variety of settings to perform a multitude of tasks such as, for instance, in assistive robotics \cite{ienca2016social} or educational robotics \cite{tanaka2014}, enhances the usefulness of improving Human-Robot Interaction (HRI) and Human-Robot Collaboration (HRC) systems \cite{ajoudani2018progress}.

Social-aware robot navigation \cite{Phani2024} and path planning algorithms become requirements for HRC in cases where robot navigation is involved. In these cases, the communication between a robot and humans can be implicit or explicit and the participants can take different roles to accomplish the task. Side-by-side navigation \cite{repiso2020ral}, human-robot handover \cite{laplaza2022} and object transportation \cite{dominguez2024} are typical cases that involve implicit or explicit communication between agents. 

However, there is a lack of HRC in most multi-agent systems that play an essential role in areas like Search and Rescue (SAR) \cite{queralta2020}, where a collaborative group of robots tries to find a target in an environment. These environments can require different types of robots like unmanned underwater vehicles (UUVs) \cite{yordanova2020}, unmanned aerial vehicles (UAVs) \cite{YANMAZ2023103018}, or unmanned ground vehicles (UGVs). Very few approaches include this collaboration in the task \cite{zheng2019collaborative, dalmasso2023shared}.

\begin{figure}[t]
    \centering
    \includegraphics[width=1\linewidth]{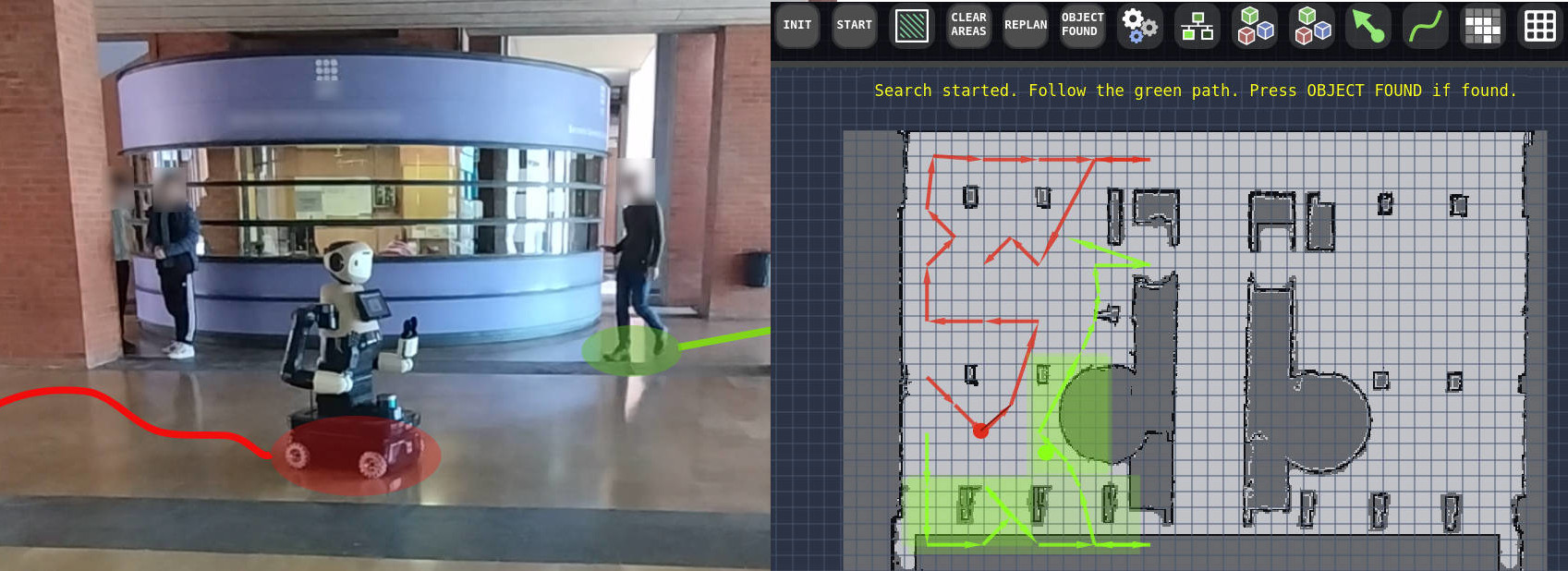}
    \caption{{\bf Human-Robot Collaborative Search with the IVO robot.} The left picture shows the IVO robot and a person searching for an object. The right picture shows the tablet computer interface with the search plans information and the human participant's preferences.}
\vspace{0mm}
\label{fig1}
\end{figure}

This work tackles the human-robot collaborative search of an object in urban environments from the point of view of Probabilistic Search (PS) and the Optimal Search Theory \cite{stone1976theory}. To achieve this end, the approach presented in this work is based on the Ant Colony Optimization (ACO) meta-heuristic, used by \cite{sara2018mmas} to solve the Minimum Time Search (MTS) problem minimizing the Expected Time (ET) to find an object. 
%Reviewer #11 punto 4, 5
The human preferences in the task are taken into account through a Human-Informed Robot Planning system to obtain plans for all agents. Different approaches to solving other tasks have demonstrated to benefit of this type of planning \cite{chen2018planning, shaikh2020measure}.
%Reviewer #14 punto 13
To simplify the problem in this work, the implementation of the algorithm supposes the same dynamical and sensorial capabilities for all agents but different preferences.

%Editor punto 1 , Reviewer 11 punto 1
The main novelties and contributions of this work are:
\begin{itemize}
    \item A new version of an MTS-ACO algorithm has been developed that takes into account learned human preferences to create search plans and allows adaptation to particular humans with a novel Human-Informed Planning using sub-priors.
    \item A Probabilistic Map Predictor based on a Convolutional Neural Network (CNN) has been developed that can learn about the most likely areas where people would look for lost objects using small datasets.  
    \item A HRC system through an interface to ensure communication between multiple agents and devices(refer to Fig. \ref{fig1}).
    \item An evaluation of a real case of HRC between a person and a robot to test the viability of the proposed methods and measure the participants' perception.
\end{itemize}

%To sum up, the novelties of this work include a new version of a MTS-ACO algorithm that takes into account human preferences, a model that learns to generate probability maps from real data and the evaluation of a real case of HRC using this system with an interface to ensure communication between agents (refer to Fig. \ref{fig1}).

The remainder of this paper is organized as follows. In Sec. II, the related work is introduced. Sec. III describes the theoretical approaches. In Sec. IV, the simulation results are presented. In Sec. V, the real-life experiment results are presented. Finally, in Sec. VI, the conclusions are provided.

\section{BACKGROUND}

In this section, the ACO meta-heuristic, which is used for the approach presented here, is briefly explained.

\subsection{The ACO Meta-heuristic}
ACO is a bio-inspired meta-heuristic that was proposed by M. Dorigo \cite{dorigo1999} to solve combinatorial NP-hard optimization problems. It has been widely used to solve the Travelling Salesman Problem (TSP), an NP-hard optimization problem. The algorithm simulates the foraging strategy of ants through a mathematical model. Ants try different routes in a graph, $G=(C, L)$, to find food in the set of nodes C and deposit pheromones, $\tau_{ij}$, each time they take an arc $(i,j)$ that connects nodes $i$ and $j$ and compounds the set L. For each ant, pheromones increase the probability of choosing the shorter arcs so, arcs with more pheromones are going to be more visited. After a while, most of the ants travel through the shorter or optimal route. Also, there is a pheromone evaporation rate, $\rho$, that allows to find new better routes.
\begin{equation}
\tau_{ij} \longleftarrow (1-\rho)\tau_{ij} \ \ \ \forall (i,j) \in L
    \label{eq1}
\end{equation}

Additionally, ACO includes a heuristic $\eta_{ij}$, to encourage the best arcs during specific optimization processes. Taking into account the pheromones and the heuristic, the probability of an ant $k$ to choose an arc $(i,j)$ is:
\begin{equation}
    p_{ij}^{k}=\frac{[\tau_{ij}]^{\alpha}[\eta_{ij}]^{\beta}}{\sum\limits_{l=1}^{C_{i}}{[\tau_{ij}]^{\alpha}[\eta_{ij}]^{\beta}}}
    \label{eq2}
\end{equation}
where $C_{i}$ is the number of available nodes from node $i$ and the parameters $\alpha$ and $\beta$ set the relative influence of the pheromones and the heuristic.

There are different ACO algorithms \cite{acobook2004} designed to improve the optimization aspects. Some approaches, like Ant System (AS), MAX-MIN Ant System (MMAS) \cite{stutzle2000max}, and Ant Colony System (ACS) have been designed to optimize in discrete spaces using graphs. Other approaches, such as the ACO for continuous domains (ACOR) can be used in continuous spaces. This work is centered on the MMAS algorithm due to the promising results obtained with UAVs \cite{sara2018mmas}.

Each approach has different equations to deposit pheromones. The rule used to deposit pheromones after the evaporation in MMAS is:
\begin{equation}
\tau_{ij} \longleftarrow \tau_{ij} + \Delta \tau_{ij}^{best}
    \label{eq3}
\end{equation}
where $\Delta \tau_{ij}^{best}=1/C_{b}$ is the pheromone applied to the arcs of the path with the minimum value for the cost function, $C_{b}$. In each optimization iteration, a group of ants produces different paths with different costs. The pheromone update can be randomly performed using the best iteration cost or the best-so-far cost.

%{\TODO {Falta definir que es Cb}} 
%{\TODO {Falta definir que es the best interaction cost}} He puesto que Cb es 'the minimum value for the cost function'

\section{RELATED WORK}

%In this section, an overview of the relevant topics related to this work is provided. Firstly, the ACO meta-heuristic is briefly explained. Secondly, there is a description of PS and Optimal Search. Finally, the main methods for Human-Robot Collaborative Search are explained.
In this section, an overview of the relevant topics related to this work is provided. These topics are a description of Probabilistic Optimal Search and different methods for Human-Robot Collaborative Search.

\subsection{PS algorithms}

PS is an area that considers probabilistic maps of the environment to find a target \cite{stone1976theory}. It has been broadly applied for military purposes and in SAR to find lost people using UAVs or UGVs. In \cite{bourgault2006optimal}, a Bayesian perspective with a greedy algorithm is proposed for maritime environments. Most models have adopted this Bayesian outlook using different optimization algorithms \cite{lanillos2012ceo, sara2016ga, sara2018mmas, sara2017acor}. The optimization utility functions used by these methods are normally the "cumulative" probability of detection or the 
%Reviewer #14 punto 2 
Expected Time (ET) to find the target. Sometimes other functions as the expended energy or the collisions are also used. When ET is the main criterion, the search is a MTS problem.

\subsection{Human-Robot Collaborative Search}

Some approaches consider HRC to find objects or people. In SAR, human-robot teams normally use these approaches to search for lost people. In \cite{williams2020collaborative}, a web interface is used to manage the search task assignment for the teams. This interface offers autonomous partitioning to assign tasks and allows the users to change it. In \cite{Papaioannou2024SynergisingHR}, UAVs search for people in disaster environments with 2 systems, one is semi-autonomous and another one is totally autonomous. In the semi-autonomous system, a Human-Informed Robot Planning is performed where the human preferences can modify the plan of the robot.

Specifically, in \cite{dalmasso2023shared}, a robot and a person share a task representation through an interface in a smartphone \cite{jenrique2021} to perform the search together. The approach uses different Social Reward Sources to enable the HRI during the task. These rewards are used to construct an objective function optimized with a Monte Carlo Tree Search planner that uses Rapidly Random Trees for each agent and it works online. This approach considers only uniform probability maps and does not consider segmented areas or the ET as a criterion.

Unlike our approach, the aforementioned methods do not combine previous knowledge about where the object could be lost, individual human preferences and the ET criterion.

\section{OUR APPROACH}
%This section describes the problem considered and the proposed approach.
\subsection{Problem Formulation} 
%This approach is focused on the human-robot collaborative search of a lost object outdoors.  In a searching problem, the unknown variable is the target state vector ${\bf x}^{t}$ (for example, its location). The prior information used to find the object is a segmented top view image where approximately equidistant nodes are sampled to build a graph $G$, taking into account the map obstacles.  The segmented image is used to obtain a prior probability map $p({\bf x}_{0}^{t})$ about the target location in a 2D map, ${\bf x}_{0}^{t}=(x_{0}^{t},y_{0}^{t})$, at the step $k=0$. 

This approach is focused on the human-robot collaborative search for a lost object outdoors. The prior information used to find the object is a segmented top-view image where approximately equidistant nodes are sampled to build a graph $G$, considering the map obstacles.
%Editor punto 4 Reviewer #14 punto 3
% Second review
To sample the nodes, the space is divided into a grid of squares. The nodes are sampled at the centroid of the squares if there is no obstacle within 40 cm, otherwise, they are sampled at the centroid of the area not covered by the obstacle in that square or at a vertex of the square. This method ensures better exploration close to obstacle edges than uniform sampling. The segmented image is used to obtain a prior probability map $p({\bf x}_{0}^{t})$ about the target location in a 2D map, ${\bf x}_{0}^{t}=(x_{0}^{t},y_{0}^{t})$, at the step $k=0$. 

During the search process, $M$ agents are only able to move in $G$, and a static target is considered so the probabilistic Markov model for the target is $p({\bf x}_{k}^{t}|{\bf x}_{k-1}^{t})=\mathbb{I}$. At each step $k$, agents can perform an observation ${\bf z}_{k}$ and the probability map is updated using the Bayes' rule and the previous observations ${\bf z}_{1:k-1}$:
\begin{equation}
p({\bf x}_{k}^{t}|{\bf z}_{1:k})=\frac{p({\bf z}_{k}|{\bf x}_{k}^{t})p({\bf x}_{k}^{t}|{\bf z}_{1:k-1})}{\int{p({\bf z}_{k}|{\bf x}_{k}^{t})p({\bf x}_{k}^{t}|{\bf z}_{1:k-1}) d {\bf x}_{k}^{t}}}
\label{eq4}
\end{equation}
where $p({\bf z}_{k}|{\bf x}_{k}^{t})$ is the observation model. To simplify the problem, a circular ideal sensor model is supposed. In this model, the probability of detecting an object in the step $k$ that is set in a location of the free space is:
\begin{equation}
p({\bf z}_{k}=D_{k}|{\bf x}_{k}^{t})=I_{A_{w}}H(R_{w}-r_{w})
\label{eq5}
\end{equation}

%Second review
where $D_{k}$ is a detection in the $k$ step, $H$ is the Heaviside function, $R_{w}$ is the considered visibility radius of the agent $w$ and $r_{w}$ is the Euclidean distance (ED) between the $w$ sensor and ${\bf x}_{k}^{t}$. $I_{A_{w}}$ is the indicator function for the agent $w$ in the not occluded area, $A_{w}$, delimited by $R_{w}$. The agent $w$ is defined as $w=\min\{1 \leq m \leq M|I_{A_{m}}(R_{m}-r_{m})>0\}$, where $r_{m}$ is the ED between the $m$ sensor and ${\bf x}_{k}^{t}$.

% Reviewer #14 punto 4
This sensor model considers obstacles or occlusions through $I_{A_{m}}$. To compute $A_{m}$, ray tracing is performed to not consider the occluded area and the obstacle area in $A_{m}$.

%This sensor model is modified when there is an obstacle or an occlusion. In case of an obstacle or occlusion ray tracing is performed to not consider the occluded area and the obstacle area. When 2 agents or more can see the target simultaneously, only the first agent is considered, so only the first term in the summation is used. This correction avoids obtaining a probability of detection higher than 1.

% Reviewer #16 punto 2
When some agents are humans, other models for object detection that combine the human field of view \cite{VONTHEIN20071065} with an estimation of how humans spin their heads while searching could be considered to obtain more realistic results. This is out of this work's scope. For this reason, a circular detection model is enough to evaluate how the approach proposed here considers human preferences in simulated and real cases.
%This sensor model is modified in intersections with obstacles where the model is zero. In field-of-view intersections, only the first term is considered. 

The goal of this approach is to find the optimal paths for the agents in the graph that minimize the ET (solve the MTS problem). This expectation is defined by:
% Reviewer #14 punto 5
\begin{equation}
    ET=\sum_{k=1}^{\infty}kp_{k}
    \label{eq6}
\end{equation}
where $p_{k}$ is the probability to find the object in the step $k$ if it has not been previously detected, ${\bf z}_{1:k-1}={\overline{D}}_{1:k-1}$, in the 2D map $S$. $p_{k}$ can be computed with $\Tilde{p}$, the unnormalized version of the probability map, as in \cite{sara2018mmas}:
\begin{equation}
    p_{k}=\int_{S}{p({\bf z}_{k}=D_{k}|{\bf x}_{k}^{t})\Tilde{p}({\bf x}_{k}^{t}|{\overline{D}}_{1:k-1}) d {\bf x}_{k}^{t}} \ \ \ \forall {\bf x}_{k}^{t} \in S
    \label{eq7}
\end{equation}
The ET computation has to be limited to a finite horizon $N$. This approximation, applied to \eqref{eq6}, does not guarantee that an optimal path will be obtained if $N$ is not enough to reduce to zero the probability of the map. For this reason, a different way to compute the ET, deduced in \cite{stone1976theory}, is used to obtain optimal paths in arbitrary horizons:
% Reviewer #14 punto 6
\begin{equation}
    ET=\sum_{k=1}^{N}[1-P(t \leq k)] \Delta t
    \label{eq8}
\end{equation}
where $P(t \leq k)=\sum_{t=1}^{k}p_{t}$ is the cumulative probability to find the object during the steps $t \leq k$ and $\Delta t$ is the time between steps, that is considered to be 1 in this formulation.

\subsection{System Overview}

\begin{figure}[t]
    \centering
    \vspace*{0.07in}
    \includegraphics[width=0.9\linewidth]{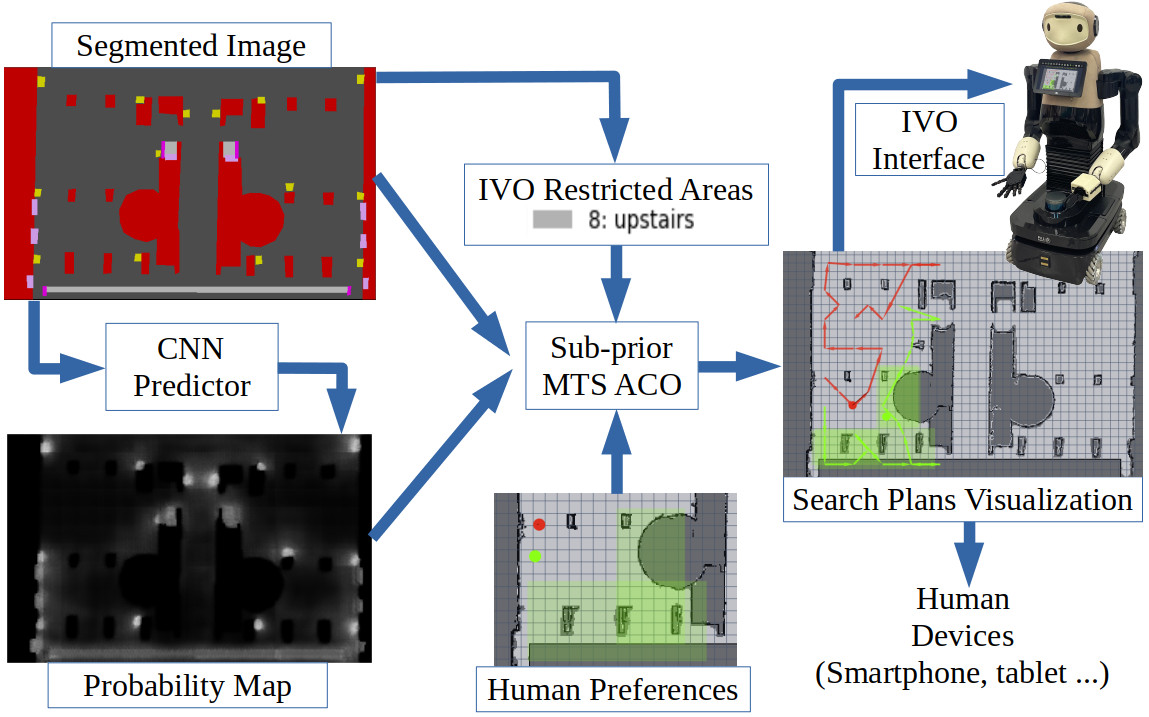}
    \caption{{\bf System-Overview.} The Segmented Map is used to predict the Probability Map and generate the restricted areas. These elements combined with the preferred areas are used to obtain the optimal paths with the Sub-Prior MTS ACO in a common representation for the IVO robot and Humans.}
\vspace{0mm}
\label{fig2}
\end{figure}

The whole centralized model is shown in Fig. \ref{fig2}. There are 2 main blocks explained in the next subsections: the Probabilistic Map Predictor and the Sub-Prior MTS-ACO algorithm. 

%Editor punto 2
The HRC is defined in this system through the next steps:
\begin{itemize}
    \item {\bf Communication of preferred areas:} The human provides the robot with the preferred areas.
    \item {\bf The robot provides the search plans:} The robot provides the plans that consider the preferred areas. 
    \item {\bf Confirmation:} The human can agree and confirm to start the search or return to the first step to get another plan.
\end{itemize}

\subsection{Probabilistic Map Predictor}

The model used to perform the prediction is a CNN with dense blocks designed in \cite{burgard2018} to predict occupancy grids. As the first step, different patches, ${\bf X_{p}}$ are obtained in a segmented bird's-eye view image, $I_{s}$, with 14 representative semantic classes in urban environments. %through a deconstruction function $F$:
%\begin{equation}
%    {\bf X_{p}} =F(I_{s})
%    \label{eq9}
%\end{equation}
During the training process, the CNN uses as ground truth a probability map $p_{GT}(\bf x_{0}^{t})$ associated with a segmented area and takes the patches as inputs:
\begin{equation}
    {\bf Y}_{p}=CNN({\bf X_{p}})
    \label{eq9}
\end{equation}
The output layer provides patches, $\bf Y_{p}$, where each pixel is the output of a spatial softmax function and represents the probability of finding a lost object. Then, the probability map, $p(\bf x_{0}^{t})$, is reconstructed using the patches, considering the average of the intersection.
\begin{figure}[t]
    \centering
    \vspace*{0.07in}
    \includegraphics[width=0.95\linewidth]{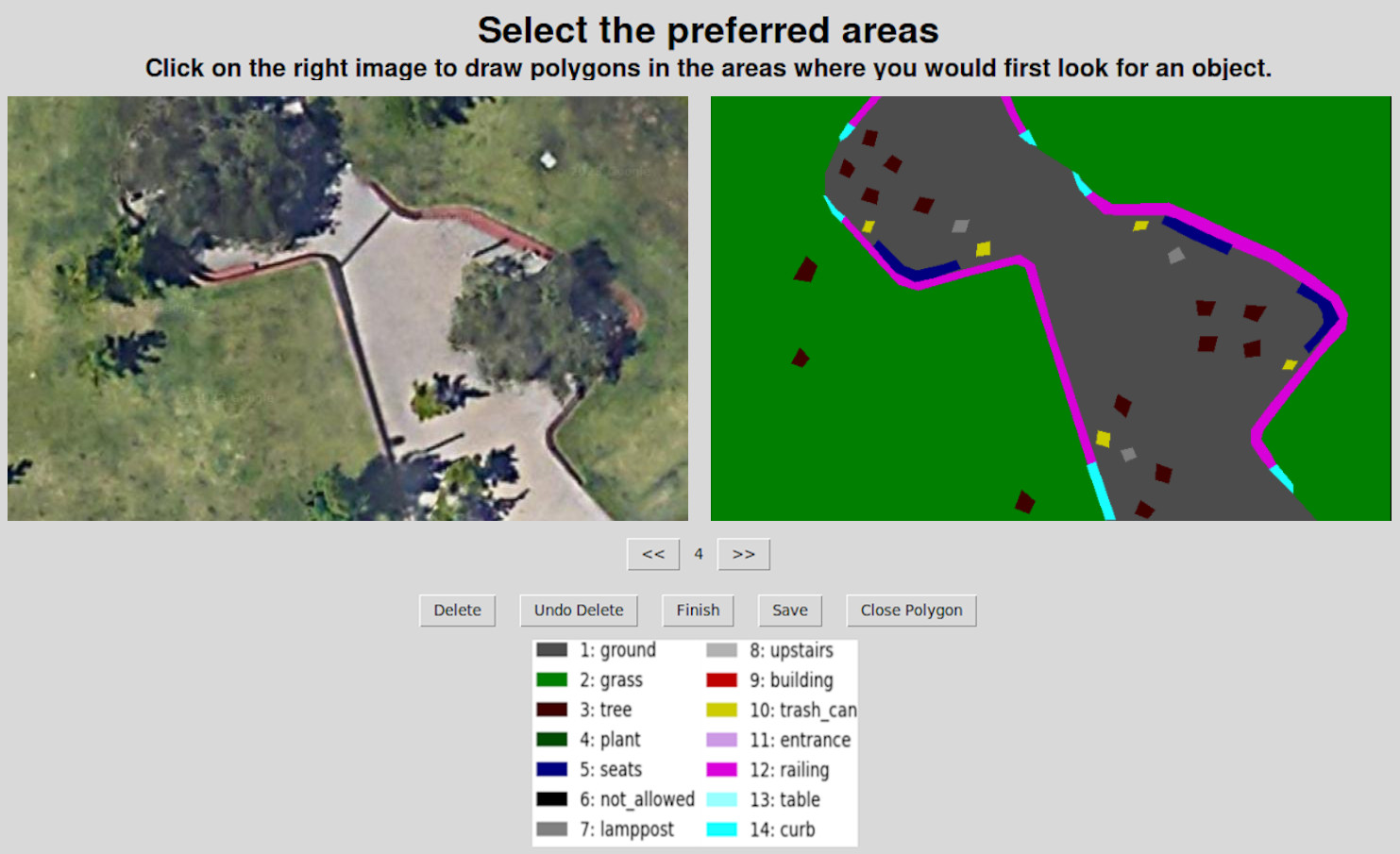}
    \caption{{\bf Labelling Interface for Users.} Participants select the areas by marking the vertices of a polygon with the computer mouse over the segmented image until the polygon is closed. The semantic classes are shown in the image.}
\vspace{0mm}
\label{fig3}
\end{figure}

To obtain a dataset to train the model, an interface using the Tkinter Python library was designed to allow 16 people to select areas in the interface for 22 segmented images where they would look for the object (refer to Fig. \ref{fig3}). %Reviewer #14 punto 8
Since the number of images is not very high, data augmentation is carried out to obtain several patches that allow to train the model and to obtain a low error in validation and testing.

% Reviewer #14 punto 9
The target considered is an object the size of a smartphone. This specification is given to the participants at the beginning and conditions the areas marked by the participants to this type of object. %The participants are encouraged to select only the first areas where they would search, without regard to specific order, number of areas, and size.
% Reviewer #14 punto 8
Participants are encouraged to select only the first areas in which they would search. They are not asked to select a specific number of areas or to do so in any order of preference. Nor are they limited in the size of the areas they select. During the process, they can also see in the interface the real top-view image corresponding to the segmented image and the 14 classes. The average of the selected areas for each map is normalized and used as $p_{GT}(\bf x_{0}^{t})$. 

\subsection{Sub-prior MTS-ACO}

This model takes $p({\bf x_{0}^{t}})$ and the segmented map as inputs to generate the agents' optimal paths. As distinct from \cite{sara2018mmas}, restricted areas are considered for each agent depending on their traversability limitations. 
%Reviewer #14 punto 11
To consider these areas, a different spatial graph for each agent is built. These graphs are constructed from the original graph (the one without restricted areas) by removing the nodes that are in the restricted areas for each agent.

% so, a different graph for each agent is considered. The nodes in the restricted areas are removed for the agent.

Another difference from ACO algorithms, like the one in \cite{sara2018mmas}, is that to consider individual preferences in the search process without losing the common objective, here it is considered a different probability map for each agent $m$. This map is called sub-prior, $p_{m}({\bf x_{0}^{t}})$. All the sub-priors are normalized to $1/M$ where $M$ is the number of agents. Then, the global probability map or prior distribution is the sum of the sub-priors.

%Reviewer #14 punto 15
\begin{figure}[t]
    \centering
    \vspace*{0.07in}
    \includegraphics[width=0.80\linewidth]{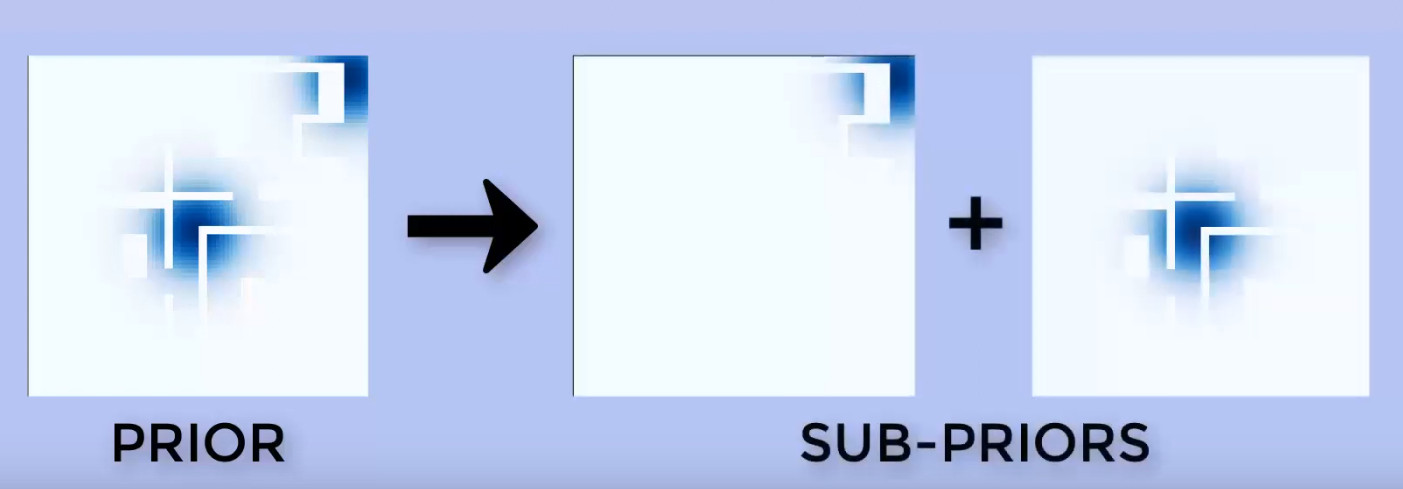}
    \caption{{\bf Gaussian sub-priors.} The left image shows a probability map of 2 gaussian functions in blue. The white lines inside the gaussians are obstacles where the probability is zero. The right images are the 2 Gaussian functions separately as sub-priors to allow the distribution of the search task between 2 agents.}
\vspace{0mm}
\label{sub_priors}
\end{figure}

In Fig. \ref{sub_priors} an example of sub-priors is presented. In this case, a prior probability map with 2 2-D Gaussian functions is divided between 2 agents into 2 sub-priors. The criteria used to perform this division depends on the negotiation process between the agents in the case of multi-agent systems with humans.

In contrast to the original ACO, 
%Reviewer #14 punto 12
as a consequence of having a different graph and a different probability map for each agent, this approach considers a different pheromone matrix for each agent, $\tau_{ij}^{m}$. This leads to the probabilities, $p_{ij}^{km}$, calculated with (\ref{eq2}), depending on the agent $m$. 
%Reviewer #14 punto 13,14
A condition is imposed in the optimization process to encourage paths of similar length in a not very restrictive way. In each step, an ant $k$ has to choose between two options: The first one is to choose a node for the agent with the shortest path. The second option is to choose a node for another random agent. The probability of choosing the first option is higher than the second one. This condition is imposed because, in the optimization process, it is supposed all the search agents have the same dynamical and sensorial features.

%A condition is imposed in the optimization process to encourage paths of similar length. In each step, an ant $k$ chooses a node for the agent with the shortest path. Another agent can be chosen with low probability to prevent this condition from being too restrictive.

Optionally, a heuristic matrix can be considered for each agent, $\eta_{ij}^{m}$, in cases where the heuristic depends on the agent. For example, it occurs when the MTS heuristic proposed in \cite{sara2018mmas} is used for this approach.

%Second review
When an ant generates the agents paths, $p_{k}$ is computed with the sub-priors for each agent, $p_{m}({\bf x}_{k}^{t}|{\overline{D}}_{1:k-1})$:
\begin{equation}
p_{k}=\sum_{m=1}^{M}\int_{S}{I_{A_{m}}H(R_{m}-r_{m})\Tilde{p}_{m}({\bf x}_{k}^{t}|{\overline{D}}_{1:k-1}) d {\bf x}_{k}^{t}}
    \label{eq10}
\end{equation}
The ET computed in this way, with the unnormalized sub-priors, $\Tilde{p}_{m}$, can be called Expected Sub-prior Time (EST) and it is normally different from the one computed in (\ref{eq8}), leading to different optimal paths in the optimization. 

%Intuitively, the sub-priors represent areas that allow to guide the formation of the agents' paths. The EST represents an ET constrained to the condition each agent visits his or her sub-prior.

%Reviewer 14 punto 15
\begin{figure}[t]
    \centering
    \vspace*{0.07in}
    \includegraphics[width=1\linewidth]{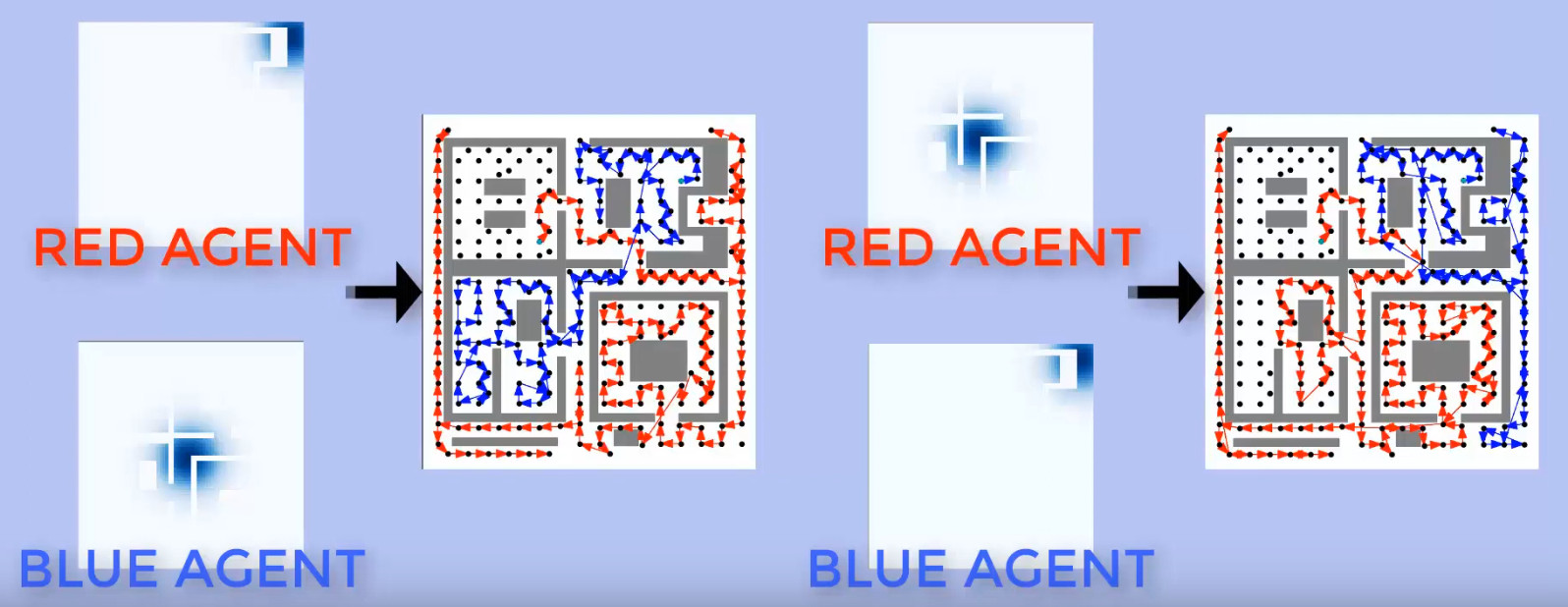}
    \caption{{\bf Sub-prior MTS-ACO planning of 2 agents exchanging sub-priors.} This figure presents the search plans for 2 agents (red and blue) in 2 different cases for the same map exchanging their sub-priors.}
\vspace{0mm}
\label{sub_priors_paths}
\end{figure}

%Reviewer #14 punto 15
Intuitively, the sub-priors represent areas that guide the formation of the agents' paths. The EST represents an ET constrained to the condition each agent visits his or her sub-prior.

In Fig. \ref{sub_priors_paths}, an example of how the sub-priors guide the generation of the paths is shown. In that figure, different search plans computed using the Sub-prior MTS-ACO (concretely, using the MMAS algorithm) are represented for 2 agents in the same map. When the upper-right gaussian is assigned to the red agent, the Sub-prior MTS-ACO generates a red path close to that area. When the central gaussian is assigned to the red agent, the red path is around that area and the blue path is around the upper-right gaussian.

%Editor punto 6 Reviewer #16 punto 8
The Sub-prior MTS-ACO has a computational complexity of  $O(IC^{2})$. The cost depends on the number of iterations, $I$, and the number of nodes, $C$. The number of agents only affects the $EST$ computation because the number of sub-priors that are updated is the same as the number of agents. For the same search map, more agents don't add a significant computational cost. On the other hand, the size of the map increases a lot the number of nodes needed and supposes an important limitation in the Computation Time (CT).

Although the algorithm supposes a static environment it can be used in environments with people or other relatively small elements that don't cause significant occlusions if the agents have capabilities for obstacle avoidance.

%\subsection{Vizanti web-based visualization}

\section{SIMULATIONS AND REAL-LIFE EXPERIMENTS}

%\subsection{CNN Predictor Results}
%The results of the predictor are shown in Table \ref{cnn_results}. The model dataset consist of 22 segmented outdoors images. 15 images are used to train the CNN, 4 images are used for validation and 3 images to test the model. 
%\begin{table}[t]
%\caption{{\bf Binary Cross Entropy (BCE) loss and Mean Squared Error (MSE) in the predicted probability map for the CNN predictor model.}}
%\label{cnn_results}
%\begin{center}
%\begin{tabular}{c c c c}
%\hline 
%Metric & Train data & Validation Data & Test Data \\
%\hline 
%{\bf BCE Loss} & 0.3902 &0.3679& 0.3065 \\

%{\bf MSE} & 0.0174 & 0.0208& 0.0132\\
%\hline
%\end{tabular}
%\end{center}
%\end

%\begin{figure}[bt]
%    \centering
%    \vspace*{0.07in}
%    \includegraphics[width=0.90\linewidth]{figures/maps2.jpg}
%    \caption{{\bf Simulation Maps.} From left to right, maps M2 and M1 are shown respectively. The third and fourth maps are the uniform and gaussian probability maps of M2. The maps size is 40 x 40 m. The agents' paths are in blue and red.} %The map without obstacles in the top right corner is M1 and the top left map is M2. The bottom left map is the M2 uniform probability map and the bottom right map is the M2 gaussian probability map. The maps size is 40 x 40 m. The agents' paths are in blue and red.}
%\vspace{0mm}
%\label{fig4}
%\end{figure}

% Reviewer #16 point 3: Figure changed 

\begin{figure}[bt]
    \centering
    %\hfill
    \vspace*{0.05in}
    \subfigure[M1 Map 40x40 m]{\includegraphics[height=0.34\linewidth]{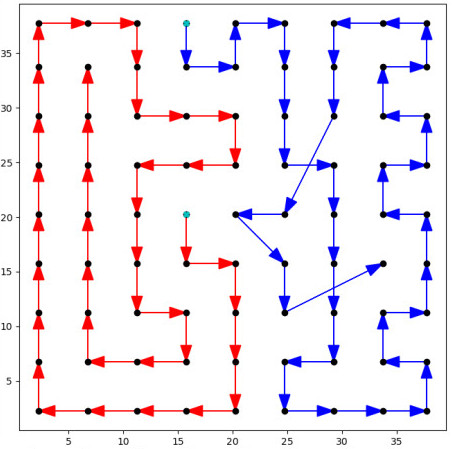}}
    %\hfill
    \subfigure[M2 Map 40x40 m]{\includegraphics[height=0.34\linewidth]{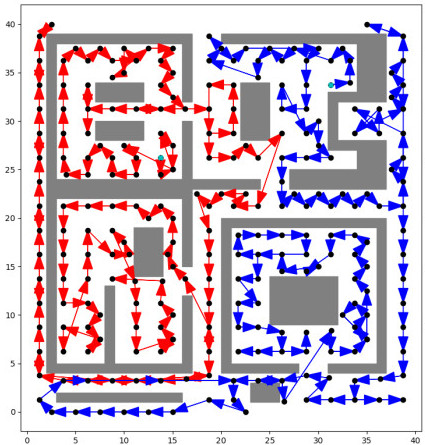}}
    %\hfill
    \subfigure[Uniform map M2]{\includegraphics[height=0.34\linewidth]{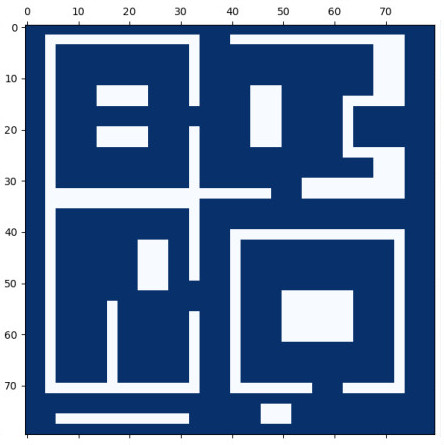}}
    %\hfill
    \subfigure[Gaussian map M2]{\includegraphics[height=0.34\linewidth]{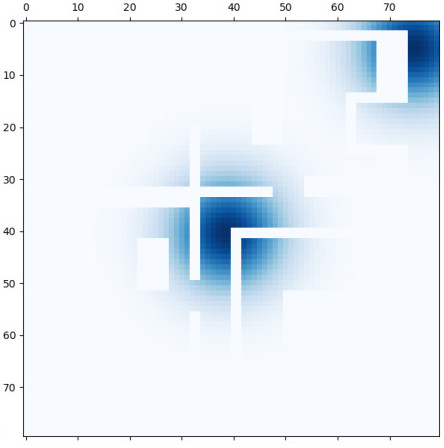}}
    %\hfill
    \caption{{\bf Simulation maps.} M1 and M2 are shown respectively in the top images from left to right. The bottom maps are the uniform and gaussian probability maps of M2. The agents' paths are in blue and red.}
    \label{fig4}
\end{figure}

\subsection{Sub-prior MTS-ACO Results}
The Sub-prior MTS-ACO model has been validated through simulations in different maps using the MMAS algorithm and 2 heuristic functions: The one used to solve the TSP problem, $\eta_{TSP}=1/d_{ij}$, that depends on the distance between nodes and the one used in \cite{sara2018mmas}, $\eta_{MTS}$, that considers regions of probability in different directions. The comparison between heuristics has been performed in 2 different maps, M1 and M2 (refer to Fig. \ref{fig4}). The metrics used to compare the heuristics are the ET, the CT, and the Path Distance (PD). The comparison is also performed considering sub-priors and non-sub-priors. 

\begin{table}[t]
\vspace*{0.05in}
\caption{{\bf ET, path distance (PD) and computation time (CT) obtained in 400 generations of 10 ants using the 2 heuristics in M1 and M2 with the uniform (U) and the gaussian (G) probability maps adding sub-priors (S).}}
\label{sim_results}
\begin{center}
\begin{tabular}{c c c c}
\hline 
\multirow{2}{*}{Map} & {\bf ET(s)} & {\bf CT(s)} & {\bf PD(m)} \\ 
 &  {\bf TSP - MTS}  &  {\bf TSP - MTS} &  {\bf TSP - MTS} \\
%Map & {\bf ET(TSP-MTS)/s} & {\bf CT(TSP-MTS)/s} & {\bf PD(TSP-MTS)/m}\\
\hline 
{\bf M1/U} & 80.98 - 94.60 & 44 - 168 & 371.42 - 839.45\\
{\bf M1/U/S} & 84.14 - 116.95 & 47 - 184 & 361.05 - 730.43  \\
{\bf M1/G} & 39.33 - 37.01 & 44 - 156 & 419.67 - 785.91\\
{\bf M1/G/S} & 38.77 - 35.15 & 48 - 184 & 416.46 - 828.24\\
\hline
{\bf M2/U} & 142.23 - 165.32 & 116 - 460 & 673.88 - 1202.69 \\
{\bf M2/U/S} & 155.47 - 237.15 & 123 - 560 & 685 - 1129.86 \\
{\bf M2/G} & 75.04 - 92.53 & 112 - 452 & 621.16 - 1098.57\\
{\bf M2/G/S} & 80.50 - 86.32 & 136 - 544 & 655.41 - 1176.21\\
\hline
\end{tabular}
\end{center}
\end{table}

The mean results of 10 trials (see Table \ref{sim_results}) show that the TSP heuristic outperforms the MTS heuristic in uniform probability maps in terms of ET. The MTS heuristic only gives slightly better results in the case where the probability is concentrated in 2 gaussians and the map has no obstacles (M1). In all the cases, the CT is much lower for the TSP heuristic because it is not necessary to compute the heuristic for each optimization step and the PD is also lower. For these reasons, the TSP heuristic is chosen to perform real-life experiments. The results in CT are shown without parallelization of the algorithm. 

The sub-priors in the uniform probability map are $p({\bf x_{0}^{t}})/2$. In the gaussian probability map, each sub-prior is a 2D gaussian assigned to the closest agent. When the sub-priors are used, the results show a small increase in ET and CT. In the case of the 2 gaussians in M1, the ET is lower because the gaussians are associated with the closest agent. If the agents' sub-priors are exchanged the ET increases a lot. These cases are sub-optimal compared to use only the MMAS due to the optimal is conditioned and the optimal EST is not exactly the optimal ET. Nevertheless, the sub-priors enable to take humans preferences into account without affect too much the performance.

\subsection{Real-life Experiments}

To check the feasibility of this approach in a real system and the ability to manage human preferences in a searching task, real-life outdoor experiments have been performed. The experiment is a searching task between our humanoid IVO robot and a person where they have to find a figure made up of 3 Parcheesi tiles in the ground, as in \cite{dalmasso2023shared}. IVO is an urban land-based robot designed to interact with humans in tasks that involve object manipulation and navigation. To navigate, IVO uses four omnidirectional wheels, a 3D-LiDAR, 2 2D-LiDAR, and a RealSense stereo camera to detect holes and ramps. IVO also possesses a touchable screen to communicate with people. 

To enable communication between IVO and the person during the search process, a web-based visualization tool for Robot Operating System (ROS) called Vizanti \footnote{Vizanti documentation: \url{http://wiki.ros.org/vizanti}} is used. This visualization is a user-friendly version of RVIZ that can be opened in a browser so, it can be used in multiple devices connected to the same local network. 

Using Vizanti and the Sub-prior MTS-ACO algorithm, a ROS implementation has been built with 3 main nodes:
\begin{itemize}
\item {\bf Search Planner:} This node computes an ordered node list using the MMAS and the sub-priors.
\item {\bf Goal Sequencer:} This node takes the node lists to generate waypoints as successive goals used for the ROS Navigation Stack.
\item {\bf Aco Gui Manager:} This node launches the Vizanti interface and communicates with the search planner node to send the user preferences, draw the paths, and manage the experiment logic.
\end{itemize}

The experiment takes around 25-30 min and consists of an initial explanation about the task and the Vizanti interface and 2 search phases in the same map: 

\begin{itemize}
\item {\bf First Search:} Here the MMAS without sub-priors is used to provide the agents' paths. The 'init' button is pressed in the interface to see the paths and 'start' is pressed to begin the search with the robot. The robot shows the same interface on its screen, so the buttons can also be pushed in the robot. The participant has to follow the green path while searching for the object. The participant can also see in the interface the red robot path, the robot position (red point), and his position (green point). When the participant finds the object has to press 'object found' to communicate it and the search finishes. If the robot finds the object or the object is not found before the robot has finished its path, a message appears to indicate the end of the search.

\item {\bf Second Search:} In this case, before the search, the participant has to press the button with a square and draw rectangles over the ROS map in the Vizanti interface. The rectangles represent the preferred areas where the person wants to search. After drawing the areas, the participant has to press 'replan' and wait until the paths computed using the MMAS with sub-priors appear. Then, after pressing 'start' the search begins.
\end{itemize}

Before starting the experiment, the task is explained and a sheet is provided with the segmented map and the semantic classes. Between the two rounds, the participant has to fill out the first part of a questionnaire and finish it at the end.

%Reviewer #16 punto 1 
%Editor punto 5
It is important to remark that the order of the search phases is the same for all participants to allow them to familiarize themselves with the interface at a basic level on the first search before using it to give their preferences. This order could induce a bias in the participants' perception that could affect the questionnaire answers. More experiments in future work are required to test whether such bias exists.

The experiment has been performed with 20 participants under the approval of the ethics committee of the Universitat Politècnica de Catalunya (UPC) \footnote{Committee website: \url{https://comite-etica.upc.edu/en}}. The volunteers are of legal age and in full use of their mental faculties. At the beginning of the experiment, they signed an informed consent form after having received the relevant information regarding the experiment. Additionally, they have accepted that all the information collected during the experiments will be treated anonymously for academic purposes.

To perform the experiment, a covered outdoor area of 21 x 27 $m$ inside the Barcelona Robot Lab has been considered. The search area is the left part of the map shown in Fig. \ref{fig1}. The robot does not use a sensor to detect the object and the person is not detected because the perception systems are not in the scope of this article. For this reason, the object position is provided to ROS and the robot finds the object when the distance to the object is closer than 2.5 $m$. The person's position during the searching task is marked by hand in Vizanti by a third person. A video with explanations about the experiment and some examples has been developed \footnote{Experiment example: \url{https://youtu.be/b0J57hXV7ic}}.

\begin{table}[t]
\vspace*{0.05in}
\caption{{\bf Average values and standard deviation in real experiments for different metrics to evaluate differences between the two search phases.}
%{\bf Average values and standard deviation in real experiments for ET, Real Search Time (RST), percent of times the robot or the person finds the object ($\%$RF and $\%$PF), percent of times the object is not found ($\%$NF), velocities of the robot or the person (${\bf\overline{v}_{r}}$ and ${\bf\overline{v}_{p}}$), Divergence Distance between the plan and the person (DD), and percent of the plan inside the Considered Areas ($\%$CA).}
}
\label{exp_results}
\begin{center}
\begin{tabular}{c c c }
\hline 
Metric & {\bf 1st search} & {\bf 2nd search} \\ 
%Map & {\bf ET(TSP-MTS)/s} & {\bf CT(TSP-MTS)/s} & {\bf PD(TSP-MTS)/m}\\
\hline 
{\bf ET ($\bf s$)} & 15.78 $\pm$ 0.00 & 20.94 $\pm$ 6.56 \\
{\bf RST ($\bf s$)} & 69.05 $\pm$ 52.40 & 86.35 $\pm$ 47.90 \\
{\bf $\%$ Robot finds} & 35 & 15 \\
{\bf $\%$ Person finds} & 50 & 75 \\
{\bf $\%$ Not found} & 15 & 10\\
${\bf\overline{v}_{r}}$ ($\bf m/s$) & 0.35 $\pm$ 0.02 & 0.33 $\pm$ 0.03 \\
${\bf\overline{v}_{p}}$ ($\bf m/s$) & 0.44 $\pm$ 0.15 & 0.51 $\pm$ 0.14 \\
{\bf DD ($\bf m$)} & 0.65 $\pm$ 0.76 & 0.46 $\pm$ 0.26 \\
{\bf $\%$ CA} & 39.95 $\pm$ 25.47 & 70.52 $\pm$ 21.51 \\
\hline
\end{tabular}
\end{center}
\end{table}

%Reviewer #16 punto 4
The results of the experiments are shown in Table \ref{exp_results}. The first metric used is the ET of the paths shown in the interface. The Real Search Time (RST) is the average time expended until the object is found or the timeout is achieved. $\%$ Robot Finds ($\%$RF) and $\%$ Person Finds ($\%$PF) are respectively the percent of times the robot and the person find the object. $\%$ Not Found ($\%$NF) is the percent of times the object is not found. ${\bf\overline{v}_{r}}$ and ${\bf\overline{v}_{p}}$ are respectively the average velocities of the robot and the person during the search task. The Divergence Distance (DD) is the average minimum distance between the plan shown in the interface and the agents' real position. The DD measures how accurately the person follows the plan displayed on the interface during the search. $\%$ Considered Areas ($\%$CA) is the average percent of the plan shown in the interface that is inside the preferred areas selected by the participants. The $\%$CA indicates the extent to which the individual's preferences are taken into account in the plans displayed in the interface.

When preferred areas are provided, the ET increases because the plan is less optimal concerning the first search, and the $\%$ CA also increases because the new plans consider human preferences. The RST is much longer than the ET because equal velocities had been considered for the agents in the optimization process with a value of 0.5 $m/s$ and the ideal sensor given by (\ref{eq5}) has been used for the agents. The results show that these assumptions are not fulfilled in this real scenario. IVO's velocity is lower than the participant's velocity for security reasons and the ideal sensor model is not a good approximation for people in real scenarios. The reduction in the DD during the second search can be explained by people's increased experience in locating themselves at the interface.

\subsection{User's Study}

%Reviewer #16 punto 1, punto 6
%Editor punto 5
A User's Study has been performed to test the next hypotheses:
\begin{itemize}
    \item {\bf H1 - }"Participants' perception of IVO changes in the second search with respect to the first one."
    
    \item {\bf H2 - }"The HRC in the planning process to obtain the search paths improves the participants' search experience."

\end{itemize}
Both hypotheses are conditioned to the fact that the order of the search phases is the same for all participants. Experiments where the order of the search is randomly taken could produce different results.
%Experiments where the order of the search is modified could produce different results.

To obtain the participants' information and test the hypotheses a questionnaire in Spanish and English has been presented with 5 sections:
\begin{itemize}
    \item {\bf Demographic Data:} In this section, the participant's name, academic level and age are taken. After the experiment, the data is anonymized. The average age of participants was 28,19 years old with a standard deviation of 4.35 years. %65 \% of participants were PhD students or have a PhD. A 20 \% were Master students or have a Master's degree and a 15 \% had a Grade or were undergraduated students.} 
    \item {\bf IVO perception after the first search:} This section evaluates the robot's perception of the participant after the first search. To evaluate the perception, questions from \cite{bartneck2009measurement} and \cite{carpinella2017TheRS} have been taken in a 7-point Likert scale to evaluate 4 attributes: Warmth, Competence, Discomfort and Anthropomorphism. 
    \item {\bf IVO perception after the second search:} This section evaluates the robot's perception of the participant after the second search. The questions are the same as in the previous section. %The average Cronbach's alpha obtained for the attributes Warmth, Competence, Discomfort and Anthropomorphism in these sections is respectively ${\bf \alpha}=(0.81, 0.61, 0.60, 0.64)$.
    \item {\bf Interface perception:} The fourth section evaluates the interface perception using the System Usability Scale (SUS) \cite{brooke1996sus}. %The obtained result, 82.06, indicates good overall usability.
    \item {\bf Preferred method:} At the end of the questionnaire, there is a last question to select one of the 2 search methods as the preferred one.
\end{itemize}

\begin{figure}[bt]
    \centering
    \vspace*{0.05in}
    \includegraphics[height=0.17\textheight]{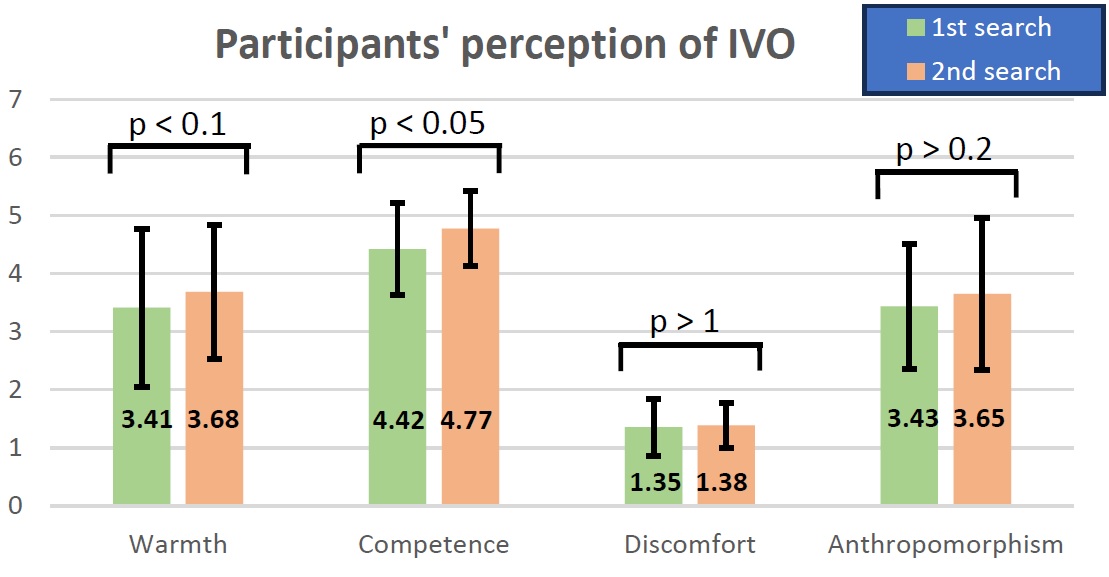}
    \caption{{\bf Participants' perception of IVO.} The mean attribute values appear in green for the first search and brown for the second search. The error bars indicate the standard deviation, and p is the p-value of the tests.}
\vspace{0mm}
\label{fig5}
\end{figure}

To test the hypotheses, the second and third section results are compared. The average Cronbach's alpha obtained for the attributes Warmth, Competence, Discomfort and Anthropomorphism in these sections is respectively ${\bf \alpha}=(0.81, 0.61, 0.60, 0.64)$. The average values of Warmth, Competence and Anthropomorphism are compared using a Paired Sample T-test. The Discomfort is the only attribute that has not passed the Shapiro-Wilk test so to compare the average Discomfort a Wilcoxon test is used. The test results are summarized in Fig. \ref{fig5}.

H1 is fulfilled for Warmth with a p-value, $p<0.1$, and Competence with $p<0.05$ and, in both cases, the second search shows a more positive participants' perception of IVO. On the other hand, for Discomfort and Anthropomorphism, the results show a very similar perception in the 2 cases.

H2 is fulfilled because of the increase in Warmth and Competence in the second search. The results in the last section of the questionnaire also support it. The 85.7 $\%$ of participants have chosen as the preferred method the one where they select the preferred areas.

The results also show that participants do not perceive a high Warmth or Anthropomorphism. These results are consistent with the fact that the person who searches does not interact too much with the robot. Most of the interaction is performed using the tablet interface. However, participants consider that the robot is very competent and the Discomfort is very low. This may occur due to the near-zero path overlapping, which prevents agents from getting in each other's way.

The interface perception has been very positive. The obtained result, 82.06, indicates good overall usability. This is consistent with the fact that the participants interact more with the interface than with IVO.

%A User's Study has been performed using a questionnaire with 4 sections. The first one is about the demographic data. The second and third parts evaluate the perception of IVO after the first and second searches in a 7-point Likert scale with some attributes used in \cite{bartneck2009measurement} and \cite{carpinella2017TheRS}. The Cronbach's alpha obtained for the attributes Warmth, Competence, Discomfort and Anthropomorphism is respectively ${\bf \alpha}=(0.81, 0.61, 0.60, 0.64)$. The Discomfort is the only attribute that has not passed the Shapiro-Wilk test so to compare the average Discomfort a Wilcoxon test is used. For other attributes, a Paired Sample T-test is performed. The results show a small increase in the Warmth and the Competence in the second search. The Discomfort is very low in all cases and the Competence is the highest rated attribute. The Anthropomorphism does not provide relevant information.   

%The last section evaluates the interface perception using the System Usability Scale (SUS) \cite{brooke1996sus}. The obtained result, 82.06, indicates good overall usability. There is a last question to select one of the 2 search methods. The 85.7 $\%$ has chosen the method using preferred areas. 

\subsection{Implementation Details}

%The MMAS parameters used for this implementation are $\alpha=1$, $\beta=6$ and $\rho=0.002$. For the experiments, a parallelized optimization is performed with 10 ants across 300 iterations, although the plans for the first search are obtained with 1200 iterations. The visibility radius considered is $2.5 \ m$, the grid distance is $3.5 \ m$ and the maximum map probability not covered by the plans is 0.014.
The MMAS parameters considered are $\alpha=1$, $\beta=6$ and $\rho=0.002$. The graph used is a grid with a 7x7 neighborhood for each cell. In the experiments' first search, plans are pre-calculated with 1200 iterations. In the second search, a parallelized optimization is performed with 10 ants across 300 iterations that take around $80 \ s$. The visibility radius is $2.5 \ m$ for all agents, the grid distance is $3.5 \ m$ and the maximum probability not covered by the plans is 0.014.

To generate probability maps, the CNN-31 model \cite{burgard2018} has been trained with the original hyper-parameters and a batch size of 32. The inputs are 14 64x64 pixel patches, one for each semantic class. The output is a 64x64 patch with the probabilities. The dataset consists of 22 segmented images close to the Barcelona Robot Lab. 15 images to train, 4 images for validation and 3 images for testing. %Reviewer #14 punto 8
Data augmentation is performed through $90^{o}$ rotations to obtain 6584 patches for training. The Mean Square Error (MSE) for training, validation and testing is respectively: 0.017, 0.021 and 0.013. The probability map used for the experiments is one of the test maps.

\section{CONCLUSIONS}

The sub-prior MTS-ACO has been presented as a feasible solution to incorporate human preferences for the search task in real experiments with a robot and a person. The results show that the algorithm can leverage the prior knowledge with the particular preferences of other agents to create optimal plans. Moreover, an interface to obtain a dataset that allows learning probability maps of the search area using a basic segmented map has been proposed. A small obtained dataset has been enough to train a CNN with a low MSE. As a third contribution, a Vizanti interface has been presented to enable HRC during the search task.

\addtolength{\textheight}{-12cm}   % This command serves to balance the column lengths

%%%%%%%%%%%%%%%%%%%%%%%%%%%%%%%%%%%%%%%%%%%%%%%%%%%%%%%%%%%%%%%%%%%%%%%%%%%%%%%%

%%%%%%%%%%%%%%%%%%%%%%%%%%%%%%%%%%%%%%%%%%%%%%%%%%%%%%%%%%%%%%%%%%%%%%%%%%%%%%%%

%%%%%%%%%%%%%%%%%%%%%%%%%%%%%%%%%%%%%%%%%%%%%%%%%%%%%%%%%%%%%%%%%%%%%%%%%%%%%%%%
%\section*{APPENDIX}


%\section*{ACKNOWLEDGMENT}

%\begin{thebibliography}{99}
\bibliographystyle{IEEEtran}
%\bibliographystyle{plain}
%\balance
\bibliography{./main}

\end{document}